\documentclass[10pt,conference]{IEEEtran}

\usepackage{graphicx}
\usepackage{booktabs}
\usepackage{amsmath}
\usepackage{amssymb}
\usepackage{url}
\usepackage{subcaption}
\usepackage{tikz}
\usetikzlibrary{positioning,shapes.geometric,arrows.meta,calc,fit}
\usepackage{multirow}
\usepackage{array}
\usepackage{colortbl}
\usepackage{xcolor}
\usepackage{float}
\usepackage{placeins}
\usepackage{stfloats}
\usepackage{balance}
\usepackage{cite}

\newcommand{\revimad}[1]{\textcolor{black}{#1}}
\newcommand{\revabd}[1]{\textcolor{black}{#1}}
\newcommand{\revrashid}[1]{\textcolor{black}{#1}}

\usepackage{algorithm}
\usepackage{algpseudocode}
\algrenewcommand\algorithmicrequire{\textbf{Input:}}
\algrenewcommand\algorithmicensure{\textbf{Output:}}
\usepackage[hidelinks]{hyperref}

\title{
ROI-Gated SAHI: Content-Adaptive Slicing-Based Inference for Efficient Object Detection
}

\renewcommand{\thefootnote}{\fnsymbol{footnote}}
\author{
Rashid Riyadh$^{1}$, Abd Ullah Khan$^{2,3}$, Imad Gohar$^{4,*}$, Muzammil Behzad$^{5,6,*}$
\\ \\
$^{1}$ Faculty of Information Technology, City University, Malaysia\\
$^{2}$ Department of Information Convergence Engineering, Kyung Hee University, South Korea\\
$^{3}$ Department of Computer Science, National University of Sciences and Technology, Pakistan\\
$^{4}$ School of Computing and Artificial Intelligence, Sunway University, Malaysia\\
$^{5}$ King Fahd University of Petroleum and Minerals, Saudi Arabia\\
$^{6}$ SDAIA-KFUPM Joint Research Center for Artificial Intelligence, Saudi Arabia
}

\begin{document}

\maketitle

\begingroup
\renewcommand{\thefootnote}{*}
\footnotetext{Corresponding author. Email: imadg@sunway.edu.my, muzammil.behzad@kfupm.edu.sa}
\endgroup

\begin{abstract}

\revrashid{Slicing-Aided Hyper Inference (SAHI) improves small object detection in high-resolution images but often spends substantial compute on background tiles. We propose \textcolor{black}{region-of-interest (ROI)}-Gated SAHI, an inference-time framework that introduces a lightweight proposer to localize foreground regions and restrict sliced refinement to informative areas. We evaluate the \textcolor{black}{framework} in two settings. On the COCO128 full split \textcolor{black}{dataset comprising 128 images, static ROI-gating is slower on average than Full SAHI, achieving a speed  ratio of 0.88, and yields a lower mAP@0.5 of 0.6602 compared with 0.7569 for Full SAHI.} A simple adaptive routing policy with $\tau =$ 0.4 \textcolor{black}{reduces the mean latency, achieving a slight gain of 1.02$\times$ over Full SAHI.} On a three-image sparse-to-dense case study, \textcolor{black}{ROI-gating achieves speedups ranging from 0.96$\times$ to 6.90$\times$ with a mean speedup of 3.41$\times$}. These results show that ROI-gating is most beneficial in sparse scenes and requires policy-based routing for robust average behavior.}

\end{abstract}

\begin{IEEEkeywords}
object detection, sliced inference, SAHI, efficiency, YOLO, edge computing.
\end{IEEEkeywords}

\section{Introduction}
\label{sec:intro}

Detecting small objects in high-resolution images is a challenging problem due to the trade-off between spatial resolution and computational efficiency~\cite{Zhu2026}. For such detection, the fine-grained spatial details must be maintained, which makes processing images at high resolutions inevitable ~\cite{NIKOUEI2025200561}. However, processing such high-resolution images using conventional object detectors is computationally expensive, leading to memory cost and inference latency~\cite{DEDE2025200505}. These constraints make it challenging to deploy object detectors in real-world particularly on resource-constrained platforms such as mobile devices, embedded systems, aerial platforms, and edge computing environments~\cite{Kong2026}.

To address this challenge, slicing-based inference has recently emerged, which improves small object detection without architectural changes or retraining. Slicing-Aided Hyper Inference (SAHI)~\cite{akyon2022sahi} is one of the slicing-based inference approaches. SAHI first partitions high-resolution images into overlapping tiles. Next, it applies object detection independently to each tile, and then merges predictions through post-processing. On one hand, by increasing the size of small objects within each slice, SAHI enhances detection accuracy. On the other hand, due to computationally efficiency, it maintains compatibility with off-the-shelf detectors. However, the computational cost of such slicing-based inference approaches increases as the number of tiles increases. As a result, substantial overhead is generated when it is uniformly applied across the entire image.

\begin{figure}[!t]
\centering
\includegraphics[width=0.5\textwidth]{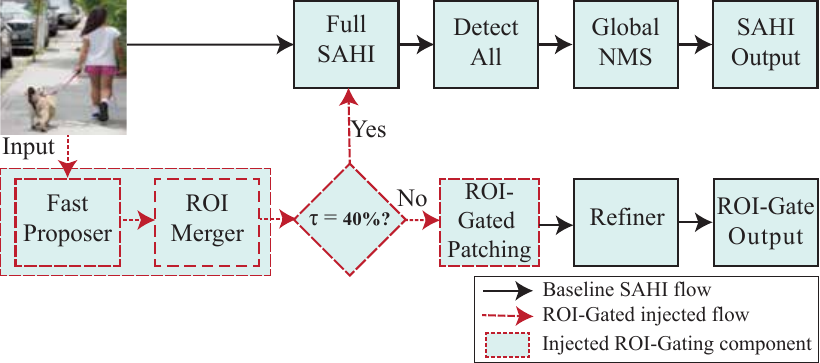}
\caption{ROI-Gated SAHI pipeline: (top) baseline full-image patching; (bottom) proposer, ROI merge, coverage check with fallback, and selective refinement. Red dashed box marks injected ROI-gating components.}
\label{fig:roi-gated-sahi}
\end{figure}

In addition to the overhead problem, another challenge associated with the existing slicing-based methods is that they perform tiled inference without discriminating the image content. As a result, background regions having little or no relevant information are processed with the same computational cost as foreground, information-rich regions. This approach becomes particularly computation-inefficient in spatially sparse or clustered objects, where a large fraction of inference time is spent on redundant computation~\cite{NEURIPS2022_2ce10f14}. This renders slicing-based approaches inapplicable for applications with latency and real-time constraints. To circumvent these challenges, inference strategies that adapt computation based on scene content are required.

Given the above challenges, an effective slicing-based technique is required that can reduce unnecessary computation in background regions while maintaining high accuracy \cite{9903420}. Second, the technique must be focused on inference time, and must not require retraining or architectural changes of detectors, thereby facilitating  deployment flexibility \cite{9943992}. Third, its performance should remain robust across diverse scene densities, including cases where foreground objects densely occupy the image \cite{pan2025review}. Finally, the technique should  be seamlessly compatible with existing slicing frameworks and detectors \cite{vogg2025computer}.

To this end, we propose a content-adaptive inference framework called region-of-interest (ROI)-Gated SAHI. The framework conditions sliced inference on foreground presence to satisfy the above mentioned requirements. For this purpose, it introduces a lightweight ROI estimation that identifies candidate foreground regions before performing the computation-intensive tiled inference process. Specifically, it avoids slicing the entire image, by performing slicing-based refinement on the selectively applied regions, i.e., regions with overlapping ROIs. This way, it bypasses background-only areas. To ensure robust behavior in dense scenes where foreground coverage is high, an intelligent fallback mechanism dynamically reverts to conventional full-image SAHI when selective gating becomes counterproductive.

The proposed framework is designed as an external inference-time system that remains fully compatible with existing detectors and slicing pipelines. By decoupling ROI reasoning from detector architectures, ROI-Gated SAHI enables adaptive computation without compromising detection completeness or requiring model retraining. This design reframes inference efficiency as a content-adaptive routing problem rather than a model-level optimization, opening a new direction for efficient deployment of slicing-based object detection systems.

The contributions are summarized as follows.

\subsection{Contributions} 
\begin{itemize}

\item We propose ROI-Gated SAHI, an inference-time framework that integrates lightweight \textcolor{black}{ROI} estimation with slicing-based inference, enabling selective activation of computationally expensive tiled refinement only in foreground-relevant regions. This design directly addresses redundant computation in background-dominated scenes while remaining fully compatible with off-the-shelf object detectors. The framework is enhanced by introducing an intelligent fallback strategy, which enables the framework to dynamically reverts to conventional full-image SAHI when the foreground size exceeds a threshold. This way, a stable and predictable performance across diverse scenes is achieved, and performance degradation in dense or cluttered scenarios is mitigated.

\item Through experimental evaluations, we exhaustively analyse the relationship between ROI coverage, tile count reduction, and inference latency using multiple image resolutions and foreground densities. We further provide insights into when ROI-gated inference should be applied for optimal results. 
Additionally, end-to-end latency, detection accuracy, and qualitative analysis of all the inference pipeline stages are provided to demonstrate the performance improvement of the proposed framework in terms of computational efficiency and detection accuracy.

\end{itemize} 
The rest of the paper is organized as follows. 
Section \ref{sec_rel_work} presents the related work. 
Section \ref{sec:method} details the proposed framework. 
Section \ref{sec:results} presents results and discussion, and Section \ref{sec:conclusion} concludes the paper.
\section{Related Work} \label{sec_rel_work}
\subsection{Small Object Detection in High-Resolution Images}
Small object detection in high-resolution images has been a crucial challenge in computer vision. Specifically, the low pixel representation, foreground–background imbalance, and large-scale spatial downsampling in convolutional networks make it challenging to detect small objects in such images \cite{NIKOUEI2025200561}. These challenges particularly impact aerial, satellite, and surveillance images, where objects of interest often occupy only a small fraction of the entire image \cite{Hua2025}. 
The existing literature deals with this problem by making  architectural enhancements~\cite{Wu2025,KOS2025110852,s25154582}. For instance, multi-scale feature pyramids is used in~\cite{9720996}, context aggregation modules in used in~\cite{YANG2026112127}, attention mechanisms is utilized in~\cite{rs14122861}, and loss re-weighting schemes is introduced in~\cite{10769544}. All these approaches improve feature representation and detection accuracy for small objects. However, they do not consider the computational cost associated with high-resolution inputs, which occurs when inference is performed densely over the entire image. Consequently, these methods cannot be efficiently applied for latency-sensitive or resource-constrained applications.

\subsection{Slicing-Based Inference for Small Object Detection}
Slicing-based inference is an effective approach to improve small object detection without changing the detector's architecture. Utilizing this idea, a SAHI framework is proposed that converts high-resolution images into overlapping tiles, applies object detection independently on each tile, and merges predictions through post-processing \cite{akyon2022sahi}. Specifically, small objects within each slice are enlarged to enable feature extraction, facilitating high accuracy gains on various detectors and datasets.

However, SAHI brings about computational overhead. For instance, the inference cost increases with increase in the number of tiles. This overhead leads to high inefficiency when the image contains sparse scenes, or when most of the image contains only background. To address this problem, adaptive SAHI (ASAHI)~\cite{rs15051249} is proposed that dynamically adjusts slice size or the number of tiles based on image resolution. Though, this approach reduces redundancy up to some extent, it uniformly applies slicing on the entire image, irrespective of foreground and background regions. As a result, the background regions still get unnecessary inference related computation. Another SAHI-based framework called $S^{3}$AHI is proposed in~\cite{10651365}, which generates high-resolution patches, and improves detection of small objects during test-time adaptation. The framework uses a teacher–student model to refine predictions. Moreover, semantic correlation graphs are constructed over sliced detections to enhance consistency. The enhanced features, however, introduce additional computational and architectural complexity to the conventional SAHI.

\subsection{ROI Mechanisms in Object Detection}
ROI methods are crucial in object detection as they focus on image regions that contain objects. In two-stage detectors, such as Faster R-CNN~\cite{7485869}, this is done using region proposal networks (RPNs), which first identifies potential object regions. These regions are then refined to produce the final detections. These methods improve detection accuracy and proposal quality. However, they have to perform high inference-time computations for high-resolution images. This is because the feature extraction process is performed over the entire image.
The latest literature explores ROI-based filtering and foreground–background separation to improve efficiency, especially in remote sensing and edge-computing applications. Techniques such as background suppression~\cite{naganuma2025}, foreground enhancement~\cite{10960542}, and ROI-based preprocessing have shown remarkable performance to reduce data transfer and processing costs. However, these methods require modifications in the detector's architecture, or retraining/task-specific tuning, which limit their applicability to the existing available models.
Contrarily, ROI-Gated SAHI introduces ROI reasoning at the inference pipeline level, which is external to the detector. Hence, it saves computations, and bypasses changes in the model's architecture or retraining.

\subsection{Positioning of ROI-Gated SAHI}
While slicing-based methods improve accuracy and ROI-based methods aim to reduce internal feature redundancy, existing approaches do not explicitly integrate lightweight foreground estimation with slicing-based inference to selectively activate expensive tiled inference \cite{muzammul2024enhancing}. ROI-Gated SAHI bridges this gap by combining (1) Lightweight ROI estimation to approximate foreground distribution at negligible cost; (2) Selective slicing, in which background-only tiles are skipped entirely; and (3) A robustness-aware fallback mechanism that reverts to full-image SAHI under dense foreground conditions \cite{li2025cn}.
This design allows ROI-Gated SAHI to achieve substantial inference-time speedups in sparse scenes while preserving detection completeness in dense scenarios \cite{suarez2025dahi}. Unlike prior work that uniformly applies slicing or requires architectural changes, our framework adapts computation dynamically to scene content and remains fully compatible with existing detectors \cite{turevckova2024artificial}.

The existing research on small object detection either improves accuracy at the expense of increased computation or reduces complexity through architectural redesign and retraining. 
However, the proposed ROI-Gated SAHI framework differs from these approaches by formulating efficiency as a content-adaptive inference problem. Specifically, it exploits foreground sparsity to gate slicing operations, aiming to reduce the redundant computation in the background-dominated regions while enhancing the accuracy benefits of slicing-based inference. This makes the proposed framework suitable for deployment in resource-constrained systems.

Motivated by this research gap, we propose a content-adaptive inference-time optimization of sliced object detection framework, called ROI-Gated SAHI, that dynamically adjusts slicing configuration according to the presence of foreground. During this operation, the framework doesn't require detector retraining or changes to the basic detector architecture. By incorporating lightweight ROI reasoning and an adaptive routing mechanism, the framework seeks to improve computational efficiency, preserve detection accuracy, and ensure practical deployment within a unified single system.

\section{Methodology}
\label{sec:method}

\subsection{Overview of ROI-Gated SAHI}

The proposed framework reduces redundant computation in the background regions by selecting foreground regions for processing. Unlike baseline SAHI, which applies uniform tiling across the entire image, our framework uses adaptive gating that restricts the processing to ROI only.
The framework consists of multiple stages, including a lightweight proposer for fast localization and a high-resolution refiner for accurate detection. To maintain steady performance in dense scenarios, an adaptive fallback mechanism is also incorporated.
Figure~\ref{fig:roi-gated-sahi} illustrates the architectural difference between baseline SAHI and the proposed ROI-Gated SAHI framework.

The proposed framework keeps the standard training procedure intact, and leverages pre-trained models from Ultralytics~\cite{yolov8_ultralytics}, initialized with COCO pre-trained weights. Both \revimad{YOLOv8n and YOLOv8s} are used without any additional training or fine-tuning. \revimad{YOLOv8n, with 3.2M parameters, serves as a fast proposer for ROI generation, while YOLOv8s, with 11.2M parameters, acts as a high-quality refiner.} The proposed ROI-gated SAHI is incorporated exclusively during the inference stage.

\subsection{Inference Pipeline}

\begin{figure*}[!t]
\centering
\includegraphics[width=\linewidth]{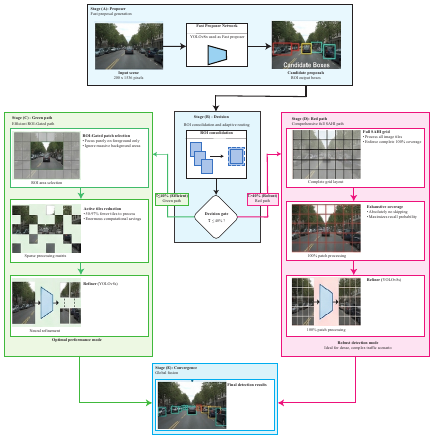}
\caption{\revimad{Complete overview of the proposed ROI-Gated SAHI framework, illustrating the sequential flow from Stage A to Stage E. The pipeline begins with high-resolution image input, followed by Stage A (Proposer), where a lightweight model generates coarse ROI candidates. Stage B computes ROI coverage and determines the routing criterion. Based on the threshold condition, the pipeline adaptively switches between Stage C (ROI-Gated SAHI), which selectively processes foreground regions, and Stage D (Full-SAHI), which processes the entire image grid. Finally, Stage E performs global detection fusion by combining outputs from the proposer and refiner branches using NMS to produce the final detections.}}
\label{fig:complete_workflow}
\end{figure*}

The complete methodology consists of five sequential stages executed during inference, as illustrated in Figure~\ref{fig:complete_workflow}.

\noindent\textbf{Stage A.} \revimad{In Stage A, referred to as the \textit{proposer stage}, the objective is to efficiently generate candidate object regions from the input image.} The pipeline begins with a high-resolution image $I \in \mathbb{R}^{H \times W \times 3}$, which is downsampled to a fixed resolution before inference, i.e.,

\begin{equation}
I' = \mathcal{D}(I), \quad I' \in \mathbb{R}^{416 \times 416 \times 3}
\end{equation}
The downsampled image $I'$ is then processed by a region proposer to outputs a set of candidate region proposals, i.e.,
 \begin{equation} \mathcal{B} = \{b_i, c_i\}_{i=1}^{n} \end{equation}
where $b_i$ denotes the predicted bounding boxes and $c_i$ represents their corresponding confidence scores. The resulting proposals provide coarse estimates of object locations with minimal computational overhead. At this stage, enhancing recall is prioritized over precision, so that as many foreground regions are captured as possible before refinement in the subsequent stages.

\noindent\textbf{Stage B.} This is the decision stage, and is used for consolidation and coverage estimation. Here, the redundant proposals are removed using Non-Maximum Suppression (NMS), i.e.,
\begin{equation}
\mathrm{IoU}(b_i, b_j) = \frac{|b_i \cap b_j|}{|b_i \cup b_j|}
\end{equation}
Boxes with $\mathrm{IoU} > 0.5$ are suppressed, yielding a refined set $\mathcal{B}^*$. Each bounding box is then expanded by a margin $\alpha = 0.15$. Consequently, 
\begin{equation}
w' = (1 + 2\alpha)w, \quad h' = (1 + 2\alpha)h .
\end{equation}
\revimad{For a box with width $w$ and height $h$, the expanded dimensions become $w' = 1.3w$ and $h' = 1.3h$, centered on the original box center. The lightweight proposer may produce slightly tight bounding boxes that crop object edges. A 15\% margin ensures objects near ROI boundaries remain fully visible to the refiner.}
As a result, adequate contextual coverage is ensured while reducing boundary artifacts. \revimad{The total ROI coverage is computed as a ratio of the aggregated ROI areas to the input image area, and is used as a metric to guide adaptive routing.}

\begin{equation}
R = \frac{\sum_{i} \text{Area}(ROI_i)}{\text{Area}(I)}
\end{equation}





\noindent\textbf{Stage C.} This stage selectively processes informative regions while reducing unnecessary computation. It consists of three main steps: (i) ROI-based patch selection focusing on foreground regions, (ii) reduced tile processing compared to Full-SAHI, and (iii) refinement using a high-capacity detector.

\revabd{If the value of $R$ is less than a specific threshold $\tau$, i.e, $R < \tau$, selective SAHI is applied only within ROI regions.} Each ROI is partitioned into overlapping tiles of size $640 \times 640$ pixels, with an overlap of $50\% - 75\%$ to preserve contextual continuity.
Let $N$ denote the total number of tiles generated in Full-SAHI over the entire image. Since only a fraction of the image is processed, the number of tiles in the ROI-Gated setting is approximated as
\begin{equation}
N_{\text{ROI}} \approx R \cdot N ,
\end{equation}
where $N_{\text{ROI}}$ represents the number of tiles processed within the ROI regions, and $R \in [0,1]$ is the ratio of total ROI area to the full image area.
Each tile is then processed using \revimad{YOLOv8s} to produce high-resolution detections. 

\noindent\textbf{Stage D.} \revimad{This stage complements Stage C, and is activated when comprehensive coverage of the image is required. Unlike the ROI-Gated approach, Full-SAHI processes the entire image without any spatial filtering.
In this setting, the full image is partitioned into a complete grid of overlapping tiles, ensuring that no region is skipped. Each tile is processed independently, enabling thorough exploration of dense and complex scenes where objects may be distributed across the entire image.
All generated patches are passed through the refiner model, \revimad{YOLOv8s}, to produce high-resolution detections.}



\noindent\textbf{Stage E.} In this stage, detections obtained from both the proposer and refiner branches are consolidated to produce the final output. The combined detection set is defined as
\begin{equation}
 \mathcal{D}_{final} = \mathrm{NMS}(\mathcal{D}_{prop} \cup \mathcal{D}_{ref}),
\end{equation}
where $\mathcal{D}_{prop}$ and $\mathcal{D}_{ref}$ denote the detections from the proposer and refiner stages, respectively. NMS is applied on the unified set with an IoU threshold of approximately $0.45$ to eliminate redundant overlapping detections and retain the most confident predictions.

\subsection{Model Latency and Analysis}
\textcolor{black}{Let $t$, $C$, and $R$ respectively represent the per-tile inference time, constant overhead, and ROI ratio, we have,}
\begin{align}
L_{\text{Full}} &= N \cdot t \\
L_{\text{ROI}} &= C + R \cdot N \cdot t
\end{align}
\revrashid{where $L_{\text{Full}}$ and $L_{\text{ROI}}$ represent latency in Full SAHI and ROI-Gated SAHI, respectively. The break-even analysis of ROI ratio defines the threshold below which ROI-gating becomes more efficient than Full SAHI in terms of latency.}
Hence,
\begin{equation}
R_{\text{break-even}} = 1 - \frac{C}{N \cdot t}
\end{equation}



\subsection{Fallback Strategy}


\revrashid{A threshold $\tau = 0.40$ is used in this stage, i.e., if $R \geq 0.40$, Full SAHI is applied; otherwise ROI-Gated SAHI is applied. The value of $\tau$ is selected from empirical latency calibration on the COCO128 full split to improve average-case robustness.}
\textcolor{black}{The selected value of $\tau$ accounts for proposer recall degradation in dense scenes, variability in computational overhead, and the effects of resolution-dependent tile distribution.} \revimad{The value of $\tau$ is empirically selected to balance computational efficiency and detection reliability.}


The proposed method reuses the same trained detectors in complementary roles, employing a fast proposer for coarse localization and a high-resolution refiner for precise detection.
\revimad{This way, computational overhead is reduced by selectively applying full inference only when necessary, while preserving detection fidelity, particularly in sparse scenes.}

\subsection{Principle of ROI-Gated Inference}
\begin{figure}[!t]
	\centering
	\includegraphics[width=1\linewidth]{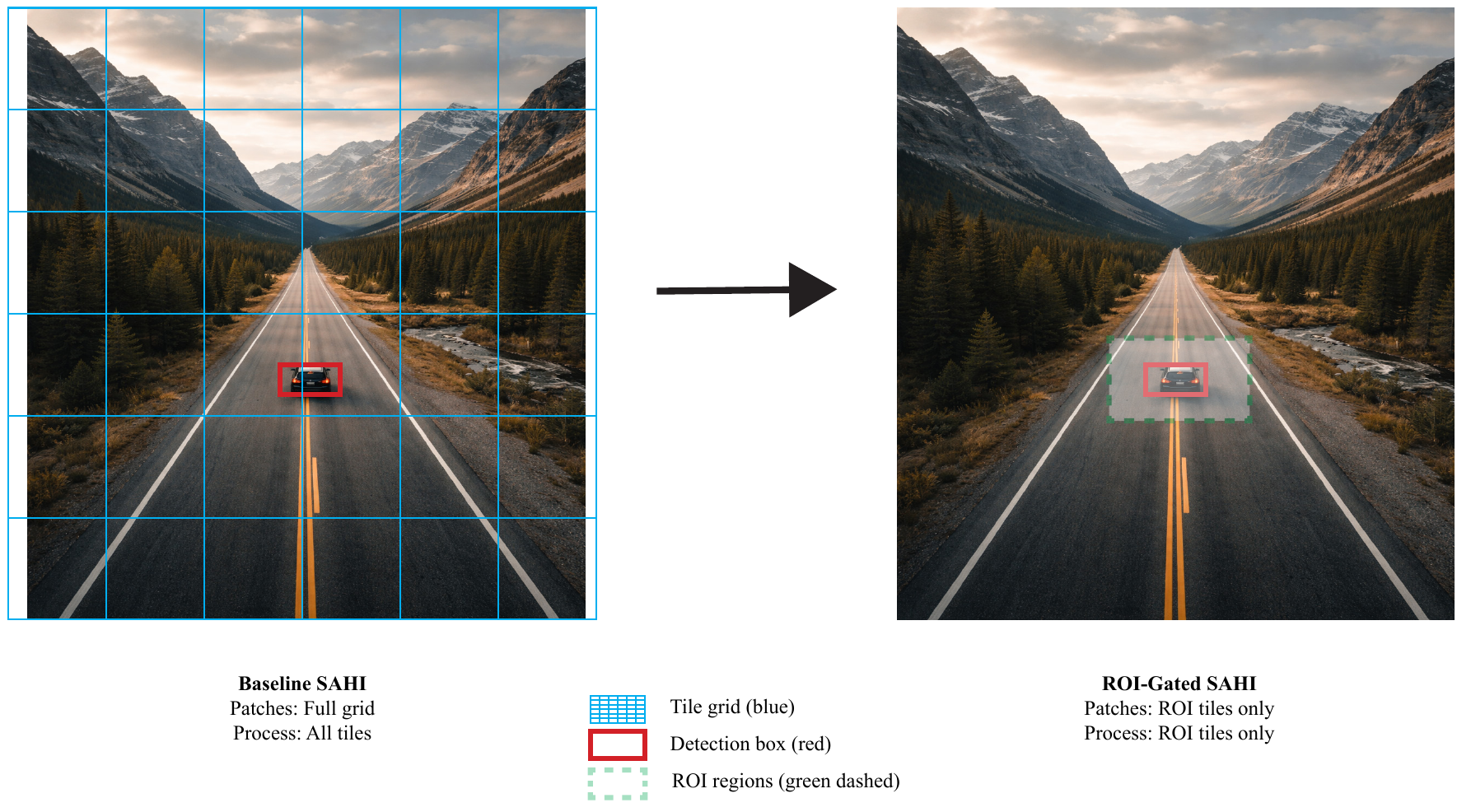}
\caption{Principle of ROI-Gated SAHI sliced inference. \textbf{Left}: Baseline SAHI processes the full image grid (blue). \textbf{Right}: ROI-Gated SAHI first identifies a candidate region (green dashed boundary) and then processes only ROI tiles. Both paths detect the same object (red rectangle) in this example, illustrating how selective tiling reduces computation cost in sparse scenes.}
\label{fig:sahi_principle}
\end{figure}

The ROI-Gated SAHI pipeline described in Section~\ref{sec:method} is utilized in the inference stage. First, the input image is processed by the proposer to identify candidate ROI. Next, the overlapping regions are merged. Depending on the ROI coverage, slicing is applied only to selected regions using SAHI. Subsequently, 
the detections obtained from the processed patches are combined using global NMS. The SAHI-based slicing is applied only during inference. Figure~\ref{fig:sahi_principle} illustrates this process, where slicing is restricted to ROI regions rather than the entire image. This way, the associated computation is reduced while keeping preserving detection rate.

 As shown in Fig.~\ref{fig:sahi_principle}, only ROI regions are processed instead of the entire image. In the left panel,  the full SAHI (baseline) divides the entire image into overlapping patches, and, as a result, all the regions are processed. In the right panel, which represents our ROI-Gated approach, a fast proposer first identifies a candidate region (green dashed boundary), and patches are generated only within that region. The detected object is shown by the red bounding box (rectangle). The figure clearly highlights the difference between processing the entire image and processing only the selected regions.

\subsection{Implementation Details}
\label{sec:implementation}
To implement the proposed framework, we developed a Python-based inference pipeline using publicly-available object detection and slicing libraries. The developed inference pipeline is reproducible across different hardware types, and is outlined in Algorithm ~\ref{alg:roi_gated_sahi}.

\begin{algorithm}[t]
\caption{ROI-Gated SAHI}
\label{alg:roi_gated_sahi}
\begin{algorithmic}[1]
\Require Input image $I$, threshold $\tau$, proposer $M_p$, refiner $M_r$
\Ensure Final detections $D_{\mathrm{final}}$

\State Generate candidate detections using the fast proposer:
\[
D_p \leftarrow M_p(I)
\]

\State Merge and expand proposal boxes to obtain ROI regions:
\[
ROI \leftarrow \mathrm{MergeExpand}(D_p)
\]

\State Compute the ROI coverage ratio:
\[
R \leftarrow \frac{\mathrm{Area}(\cup_i ROI_i)}{\mathrm{Area}(I)}
\]

\If{$R < \tau$}
    \State Generate SAHI tiles only from ROI-overlapping regions:
    \[
    T \leftarrow \mathrm{ROI\text{-}GatedSAHI}(I, ROI)
    \]
\Else
    \State Generate SAHI tiles over the full image:
    \[
    T \leftarrow \mathrm{FullSAHI}(I)
    \]
\EndIf

\State Refine selected tiles using the high-resolution detector:
\[
D_r \leftarrow M_r(T)
\]

\State Fuse proposer and refiner detections:
\[
D_{\mathrm{final}} \leftarrow \mathrm{NMS}(D_p \cup D_r)
\]

\State \Return $D_{\mathrm{final}}$
\end{algorithmic}
\end{algorithm}

As this work focuses on inference-time optimization, we utilize YOLOv8n and YOLOv8s from Ultralytics~\cite{yolov8_ultralytics} as the proposer and refiner, respectively. Both of the models were initialized with weights pre-trained on the COCO dataset. PyTorch is used for fast inferential computations, while NumPy and OpenCV are used for image preprocessing and bounding box operations. Detection accuracy and execution time were used as performance indicators for all experiments. 

\subsection{Evaluation Strategy}
Two types of evaluation settings are adopted to ensure a comprehensive evaluation. The first one is a dataset-based evaluation and the second one is a case-study-based evaluation. 
The former is conducted on the full COCO128 dataset. This dataset contains on 128 images. This evaluation highlights the overall system performance. The second evaluation is based on three high-resolution images representing sparse, moderate-dense (called moderate for brevity), and full-dense (called dense for brevity) object distributions. This configuration is essential to carry out a detailed analysis of latency and detection behavior under diverse spatial environments. These two evaluation settings are reported separately to avoid conflating aggregate dataset-level statistics with localized behavior observed in specific scene configurations.

\section{Results and Discussion}
\label{sec:results}

\subsection{Quantitative Results}
The proposed method is evaluated on the COCO128 test set and compared against standard SAHI-based inference. The comparison is conducted from three complementary perspectives, namely overall detection accuracy, inference efficiency in terms of processing speed, and robustness under varying object sparsity conditions in high-resolution images. To provide a more fine-grained analysis, we further consider three representative scenarios corresponding to dense, moderate, and sparse object distributions. These cases are used to systematically assess the behavior of the proposed approach under different spatial complexity levels.

\subsection{Overall Accuracy}

\begin{figure}[!htbp]
\centering
\includegraphics[width=0.48\textwidth]{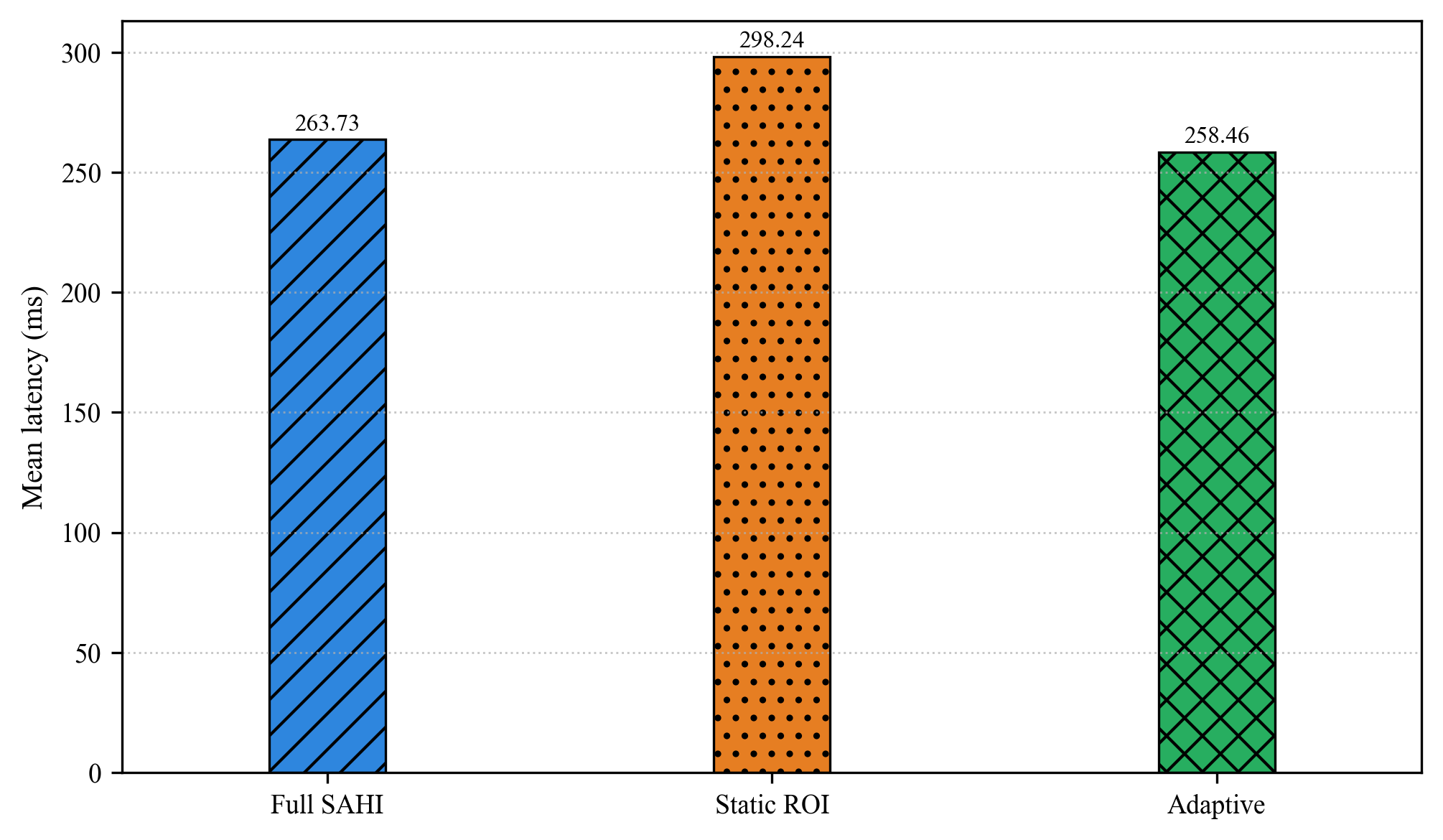}
\caption{Latency Comparison: average latency for Full SAHI, Static ROI-Gated SAHI, and Adaptive routing (threshold 0.40).
This figure compares the average runtime of the three options on the full dataset. Static ROI-Gated is slower overall, while Adaptive routing is the fastest on average because it uses ROI only when ROI area is small.}
\label{fig:coco128_core_overall_latency}
\end{figure}

Table~\ref{tab:unified_results} shows that the static ROI-Gated SAHI configuration is overall slower than Full SAHI, with a speed ratio of 0.88 (298.24~ms vs.\ 263.73~ms) also shown in Figure~\ref{fig:coco128_core_overall_latency}, while achieving lower detection accuracy in terms of mAP@0.5 (0.6602 vs.\ 0.7569). The observed latency improvement is limited to sparse scenes ($R < 0.30$), whereas both moderate and dense regimes consistently exhibit higher inference time compared to Full SAHI on average. In contrast, the adaptive routing strategy with $\tau = 0.40$ improves overall efficiency, reducing the average latency to 258.46~ms as shown in Figure~\ref{fig:coco128_core_overall_latency}, corresponding to a $1.02\times$ speedup relative to Full SAHI, by selectively routing 26 out of 128 images to the ROI-based processing branch.

\begin{figure}[!htbp]
\centering
\includegraphics[width=0.48\textwidth]{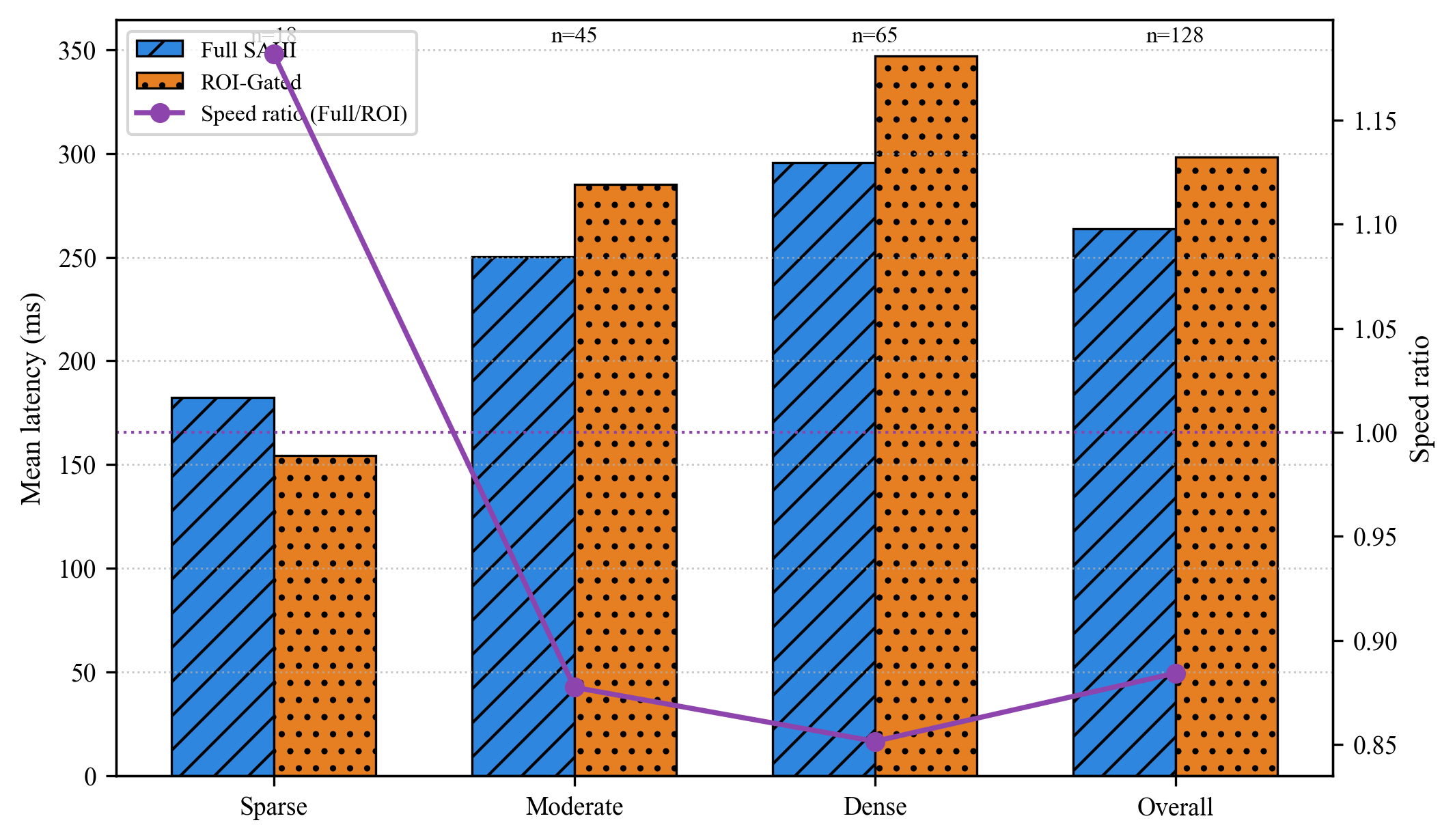}
\caption{COCO128 full split (128 images): adaptive threshold sweep showing mean adaptive latency and number of images routed to ROI.
This figure shows how performance changes when we move the routing threshold. As threshold changes, the number of images sent to ROI changes, and latency follows a curve. The paper setting (0.40) gives near-best latency while keeping routing behavior stable.}
\label{fig:coco128_core_regime_behavior}
\end{figure}

Figure~\ref{fig:coco128_core_regime_behavior} presents a regime-wise comparison of mean latency between the Full SAHI and ROI-Gated approaches across sparse, moderate, and dense subsets, as well as the overall dataset, with the corresponding speed ratio (Full/ROI) overlaid. In the sparse regime, ROI-Gated processing achieves a clear latency advantage, yielding a speed ratio greater than 1.0 and demonstrating the benefit of selectively processing only limited regions of interest. However, as the scene complexity increases, this advantage diminishes: in the moderate and dense regimes, the ROI-Gated approach incurs higher latency than Full SAHI, reflected by speed ratios below 1.0. This reason for this degradation is the increased number of regions required to be processed. Processing such a higher number of regions gives rise to routing and coordination overheads. The performance gap is more visible in dense scenes. 
This shows that ROI-based approach doesn't perform better when the majority area inside the image contains relevant objects. The overall results also show this fact, i.e., on average, ROI-Gated processing slightly performs lower than Full SAHI. The results imply that ROI-based routing is highly efficient in sparse scenes. Also, they show that ROI-based routing brings high gains in sparse regions but becomes less advantageous as image density increases.

\begin{figure}[!htbp]
\centering
\includegraphics[width=0.48\textwidth]{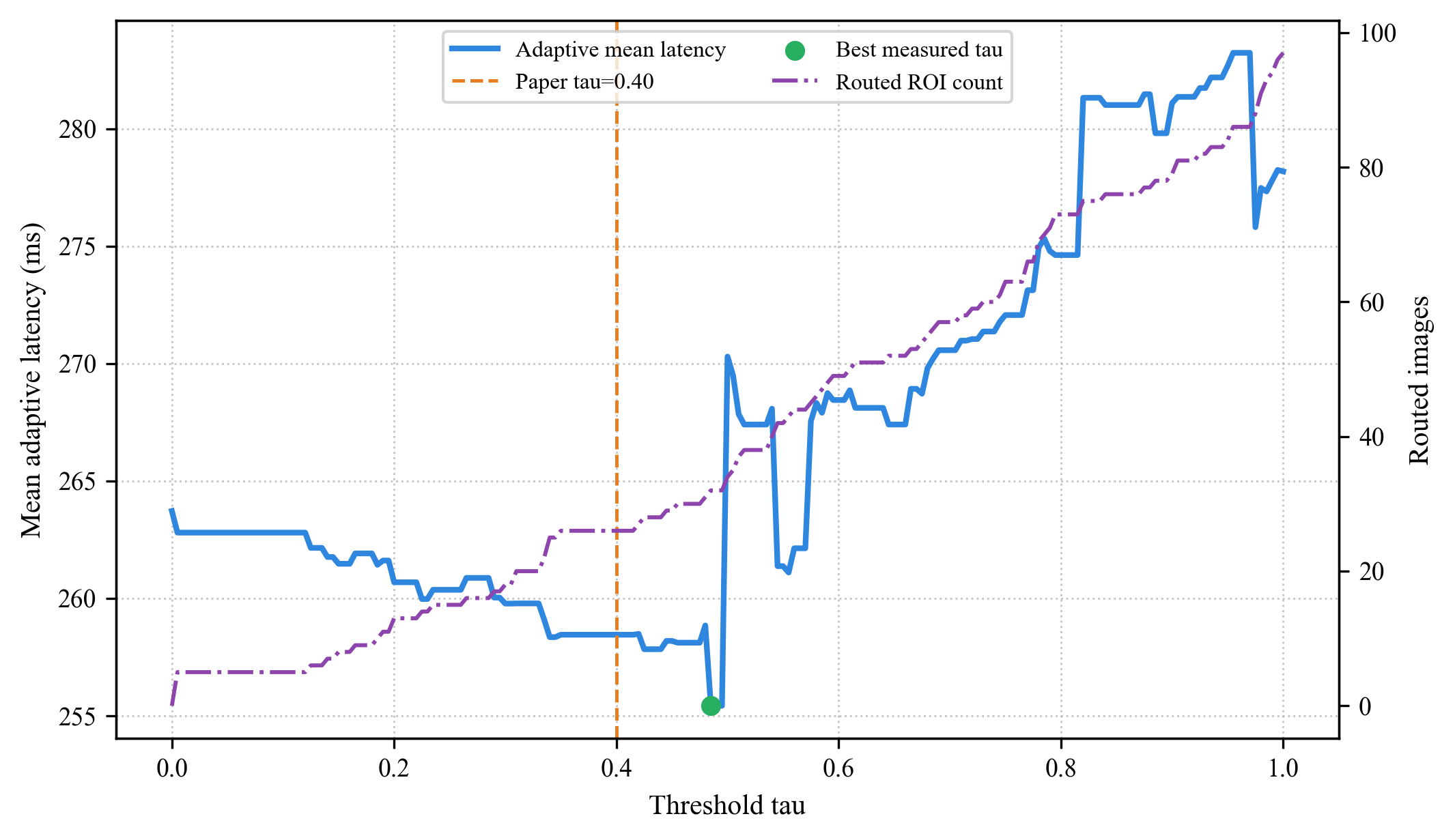}
\caption{Thresholding for $\tau$: adaptive threshold sweep showing mean adaptive latency and number of images routed to ROI.
This figure shows how performance changes when we move the routing threshold. As threshold changes, the number of images sent to ROI changes, and latency follows a curve. The paper setting (0.40) gives near-best latency while keeping routing behavior stable.}
\label{fig:coco128_core_threshold_sweep}
\end{figure}

Figure~\ref{fig:coco128_core_threshold_sweep} shows the system performance as a function of the routing threshold $\tau$. Both the mean adaptive latency and the number of images routed to the ROI pipeline are shown in the figure. As can be seen, as $\tau$ increases, the number of images for ROI processing increases, also increasing the number of routed images. 
Aside from this, the mean latency initially decreases because low-confidence regions are processed in a selective fashion. This decrease touches a minimum around $\tau \approx 0.5$. Thereafter, the mean latency increases because the increased number of routing introduces extra computational overhead. 
The results shows that trade-off exists between selective processing and routing cost. The small value of threshold limits efficiency gains, while its high value increases efficiency. Our selected operating point ($\tau=0.40$) is near the optimal region. 
This operating point achieves near-minimal latency. It also maintains stable routing behavior. As a result, it provides a good balance between performance and computational cost.

\begin{figure}[!htbp]
\centering
\includegraphics[width=0.48\textwidth]{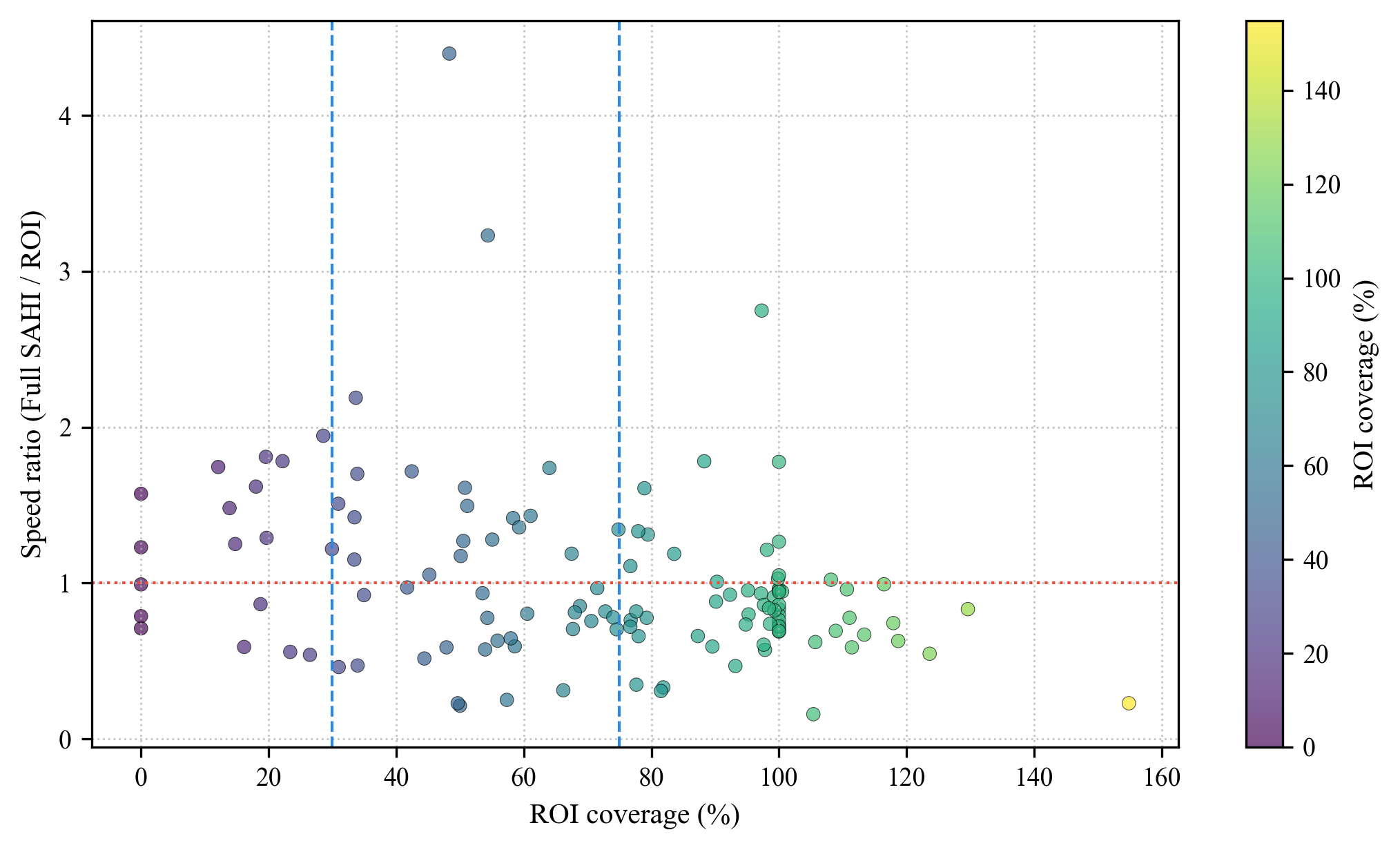}
\caption{Speed Ratio: per-image speed ratio versus ROI coverage, with break-even and regime boundaries.
Each point is one image. Points above 1.0 mean ROI is faster; below 1.0 mean Full SAHI is faster. The plot shows that ROI tends to help more when ROI coverage is low, and tends to lose advantage as ROI coverage increases.}
\label{fig:coco128_core_roi_speed_scatter}
\end{figure}

Figure~\ref{fig:coco128_core_roi_speed_scatter} shows how the per-image speed ratio (Full SAHI / ROI) changes as a function of  ROI coverage. Each dot/point represents an  image. The horizontal dotted line at the speed ratio of 1 represents the break-even point. The value above this point indicates that ROI-based approach is more efficient while values below this point indicates that full SAHI processing is more efficient.  The vertical dotted lines divide the data into low-, medium-, and high-coverage regions. 
As can be noted from the figure, 
for images at low ROI coverage, most points lie above the break-even line. This implies that ROI-based approach yields consistent speed improvements. This is because, ROI-based approach restricts the required computations to a small subset of the images. 
As ROI coverage increases, the acquired speed improvement becomes smaller. 
In the medium coverage regions, the results are more diverse. In this region, the points are distributed on both sides of the break-even line.

In the high coverage region, most points lie below the break-even line. This implies that  the extra overheads resulting from ROI extraction and routing exceeds its benefits, which makes full SAHI more faster and efficient. Overall, these results suggest taht the ROI-GAted SAHI is more effective in sparse scenes. Its effectiveness decreases as ROI coverage increases, i.e., when a large part of the image requires detailed processing. In such a case, the advantages of selective routing decreases.

\begin{table*}[!t]
\centering
\caption{Quantitative evaluation on COCO128. Speed is reported as mean Full/ROI latency ($>1$ favors ROI-Gated). ROI Faster Count reflects per-image gains (Adaptive: routing). $R=\{0.30, 0.75\}$, $\tau=0.40$. F1 and Mean IoU report agreement with Full SAHI.}
\label{tab:unified_results}
\scriptsize
\setlength{\tabcolsep}{3pt}
\resizebox{\textwidth}{!}{%
\begin{tabular}{lcccccccccc}
\toprule
\textbf{Setting} & \textbf{Images} & \textbf{ROI (\%)} & \textbf{Full mAP@0.5} & \textbf{ROI mAP@0.5} & \textbf{Full Lat. (ms)} & \textbf{ROI Lat. (ms)} & \textbf{Speed} & \textbf{ROI Faster Count} & \textbf{Agreement F1} & \textbf{Agreement Mean IoU} \\
\midrule
Overall & 128 & 68.79 & 0.7569 & 0.6602 & 263.73 & 298.24 & 0.88 & 45/128 & 0.7828 & 0.9100 \\
Sparse regime ($R<0.30$) & 18 & $<30$ & 0.6459 & 0.3630 & 182.42 & 154.38 & 1.18 & 11/18 & 0.6981 & 0.7856 \\
Moderate regime ($0.30 \leq R < 0.75$) & 45 & 30--75 & 0.7265 & 0.5605 & 250.27 & 285.22 & 0.88 & 20/45 & 0.6727 & 0.9108 \\
Dense regime ($R \geq 0.75$) & 65 & $>75$ & 0.8063 & 0.7712 & 295.56 & 347.09 & 0.85 & 14/65 & 0.8824 & 0.9439 \\
Adaptive policy ($\tau=0.40$) & 128 & dynamic & 0.7569 & 0.7305 & 263.73 & 258.46 & 1.02 & routed 26/128 & 0.9381 & 0.9638 \\
\bottomrule
\end{tabular}
}
\\[2pt]
{\footnotesize\textit{Note: Full SAHI values are shown in the Full mAP@0.5 and Full Lat. columns. Agreement metrics (F1, Mean IoU) compare ROI-Gated outputs against Full SAHI outputs. In the Adaptive row, ROI mAP@0.5 and agreement metrics are computed for the mixed routing policy output.}}
\end{table*}

\subsection{Performance Dynamics Across ROI Regimes}

To evaluate the efficiency of the proposed method under varying scene complexities, we conducted an analysis on three representative high-resolution images from the COCO128 dataset. These images were selected to represent three distinct operating regimes: \textbf{sparse}, \textbf{moderate}, and \textbf{dense} Region of Interest (ROI) coverage. We quantify the performance using latency, speedup, ROI coverage, and detection agreement as summarized in Tables \ref{tab:frames}, \ref{tab:summary}, \ref{tab:per_image_speed} and \ref{tab:per_image_agreement}.

\begin{table}[t]
\centering
\caption{\revrashid{Three-image case-study frame statistics and detection counts.}}
\label{tab:frames}
\scriptsize
\begin{tabular}{c|p{1cm}|p{.8cm}|p{0.8cm}|c|p{0.75cm}|p{0.5cm}}
\toprule
\textbf{Image} & \textbf{Resolution} & \textbf{Full} & \textbf{ROI} & \textbf{Speed} & \textbf{Speedup} & \textbf{ROI\%} \\
 & & \textbf{detection} & \textbf{detection} & \textbf{(ms)} & \textbf{($\times$)} & \\
\midrule
image1 & 1200×800 & 102 & 72 & 341/355 & 0.96 & 69.9 \\
image2 & 1536×1024 & 13 & 4 & 579/244 & 2.38 & 26.4 \\
image3 & 1024×1536 & 2 & 1 & 527/76 & 6.90 & 2.7 \\
\bottomrule
\end{tabular}
\end{table}
\begin{table}[t]
\centering
\caption{\revrashid{Aggregate performance summary for the three-image case study.}}
\label{tab:summary}
\small
\begin{tabular}{l|r}
\toprule
\textbf{Metric} & \textbf{Value} \\
\midrule
Frames Tested & 3 \\
Mean Speedup & 3.41$\times$ \\
Mean ROI Coverage & 33.0\% \\
Total Full SAHI Detections & 117 \\
Total ROI-Gated Detections & 77 \\
Detection Agreement (Mean F1) & 0.544 \\
\bottomrule
\end{tabular}
\end{table}

\subsubsection{Latency and Speedup Analysis}

\begin{figure}[!htbp]
\centering
\includegraphics[width=0.48\textwidth]{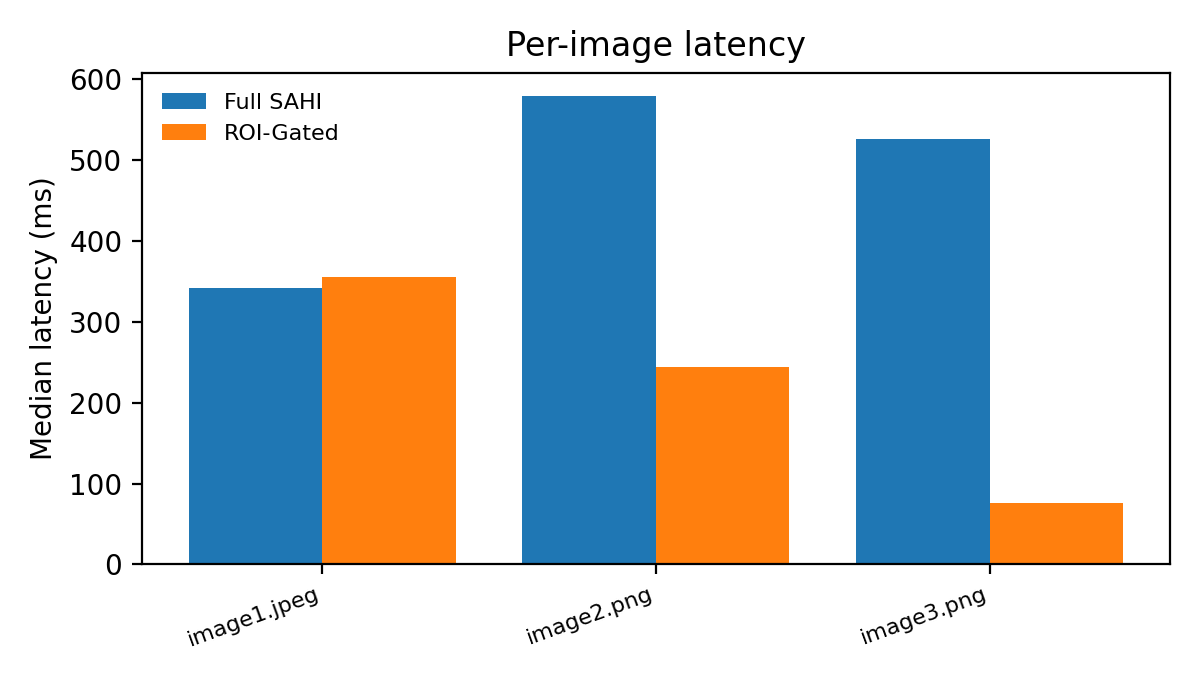}
\caption{Per-image latency comparison between Full SAHI and ROI-Gated SAHI.}
\label{fig:latency}
\end{figure}

The evaluation demonstrates that ROI-Gated SAHI achieves a mean speedup of 3.41$\times$ across the selected test cases with an average ROI coverage of 33.0\% (see Table \ref{tab:summary}). As illustrated in Figure \ref{fig:latency}, the proposed method significantly reduces inference latency in sparse and moderate scenes. 

For the sparse scene (\textit{image3.jpeg}), where the ROI covers only 2.7\% of the image, latency drops from 526.52 ms to 76.25 ms, yielding a 6.90$\times$ speedup. In the moderate scene (\textit{image2.jpeg}) with 26.4\% ROI coverage, latency is reduced by more than half, from 579.01 ms to 243.69 ms (2.38$\times$ speedup). Conversely, in the dense scene (\textit{image1.jpeg}) where ROI coverage reaches 69.9\%, the overhead of the gating mechanism results in latency comparable to the baseline (0.96$\times$ speedup).

\begin{table}[t]
\caption{\revrashid{Per-image latency and ROI coverage.}}
\label{tab:per_image_speed}
\centering
\scriptsize
\begin{tabular}{lrrrc}
\toprule
Image & Full latency & ROI latency & Speed & ROI \\
 & (ms) & (ms) & ($\times$) & (\%) \\
\midrule
img1.jpeg & 342 & 355 & 0.96 & 69.9 \\
img2.jpeg & 579 & 244 & 2.38 & 26.4 \\
img3.jpeg & 526.52 & 76.25 & 6.90 & 2.7 \\
\bottomrule
\end{tabular}
\end{table}

\begin{figure}[!htbp]
\centering
\includegraphics[width=0.48\textwidth]{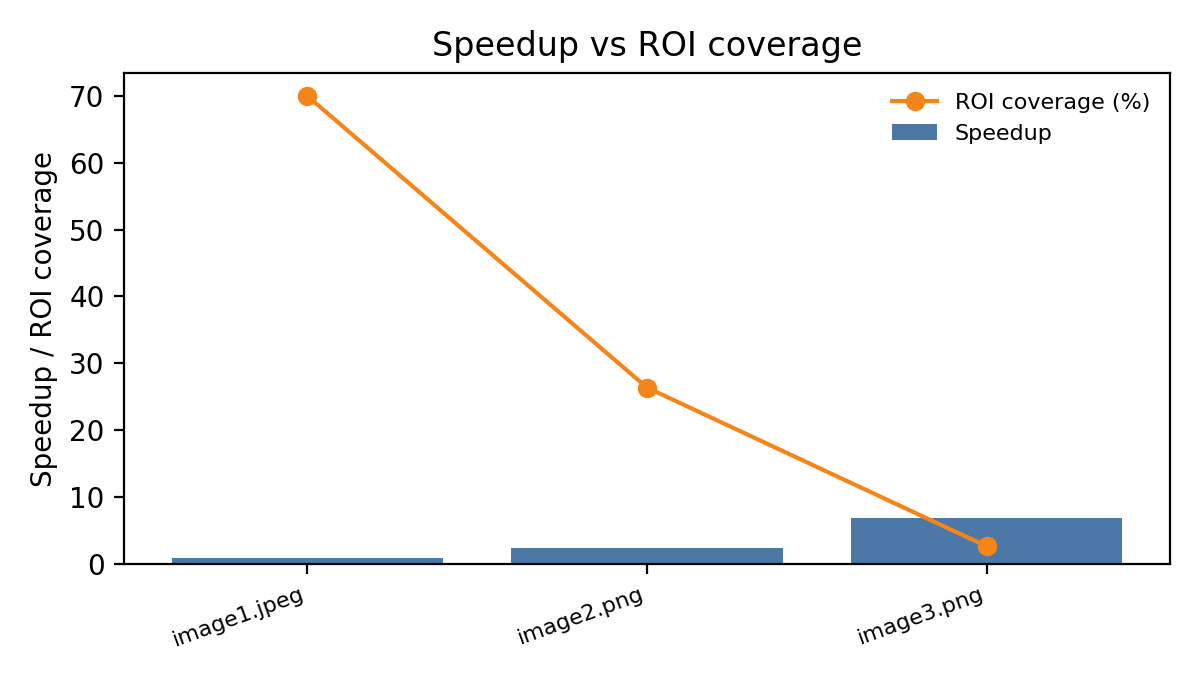} 
\caption{(Speedup vs ROI Coverage) Per-image speedup across different ROI regimes.}
\label{fig:speedup_bars}
\end{figure}

The relationship between foreground density and efficiency is further explored in Figure \ref{fig:speedup_bars}. The data confirms that speedup is inversely proportional to ROI coverage; maximum computational gains are achieved when the target objects occupy a small portion of the total image area, thereby allowing the system to skip a larger number of tiles.

\subsubsection{Detection Agreement and Robustness}

To ensure that computational efficiency does not come at the cost of detection quality, we measured the agreement between ROI-Gated SAHI and the Full SAHI baseline. As shown in Table \ref{tab:per_image_agreement}, while the F1 scores reflect differences in total detection counts (mean F1 of 0.544), the spatial alignment of the resulting bounding boxes remains exceptionally high. 

The Mean IoU ranges from 0.87 to 0.98, indicating that when objects are detected, their localization is highly consistent with the full-inference baseline. This suggests that the ROI-Gating mechanism effectively preserves spatial accuracy while successfully discarding redundant background processing.

\begin{table}[t]
\caption{Agreement of ROI-gated SAHI vs. full-image SAHI (IoU matching).}
\label{tab:per_image_agreement}
\centering
\begin{tabular}{lrrrr}
\toprule
Image & F1 & Precision & Recall & Mean IoU \\
\midrule
image1.jpeg & 0.49 & 0.60 & 0.42 & 0.89 \\
image2.png & 0.47 & 1.00 & 0.31 & 0.87 \\
image3.png & 0.67 & 1.00 & 0.50 & 0.98 \\
\bottomrule
\end{tabular}
\end{table}

\subsection{Qualitative Results}
\begin{figure*}[!t]
    \centering
    \begin{subfigure}[b]{0.19\textwidth}
        \centering
        \includegraphics[width=\linewidth]{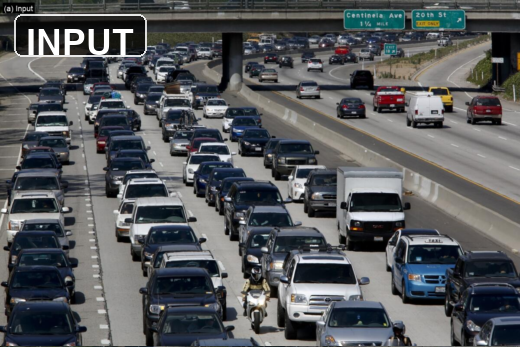}
        \caption{Input}
        \label{fig:qual7_input}
    \end{subfigure}
    \hfill
    \begin{subfigure}[b]{0.19\textwidth}
        \centering
        \includegraphics[width=\linewidth]{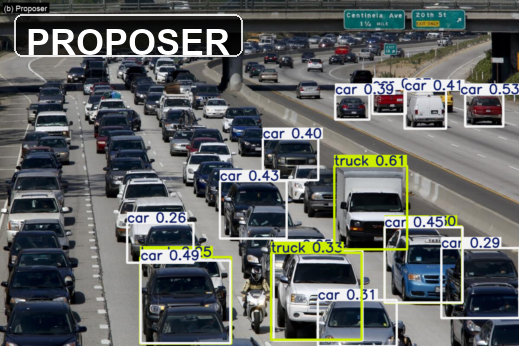}
        \caption{Proposer Detections}
        \label{fig:qual7_proposer}
    \end{subfigure}
    \hfill
    \begin{subfigure}[b]{0.19\textwidth}
        \centering
        \includegraphics[width=\linewidth]{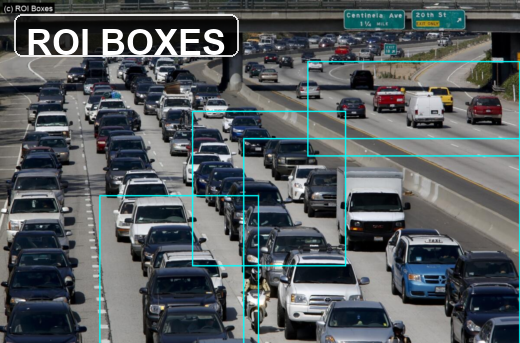}
        \caption{Merged ROI Regions}
        \label{fig:qual7_roi}
    \end{subfigure}
\hfill
    \begin{subfigure}[b]{0.19\textwidth}
        \centering
        \includegraphics[width=\linewidth]{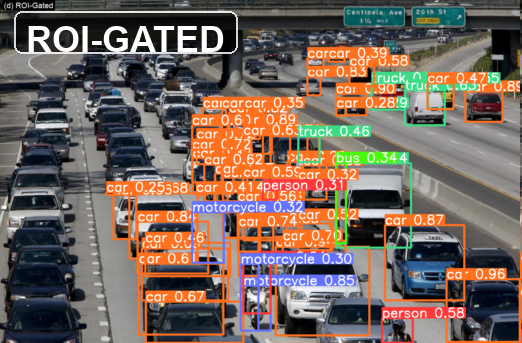}
        \caption{ROI-Gated SAHI}
        \label{fig:qual7_roi_gated}
    \end{subfigure}
    \hfill
    \begin{subfigure}[b]{0.19\textwidth}
        \centering
        \includegraphics[width=\linewidth]{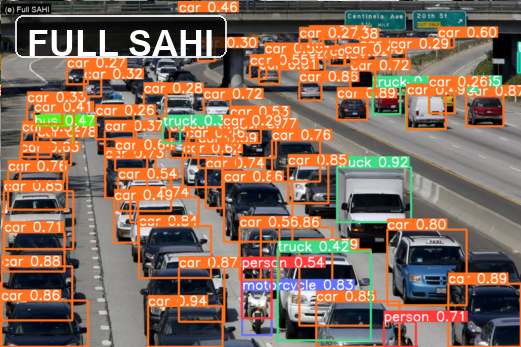}
        \caption{Full SAHI Baseline}
        \label{fig:qual7_full_sahi}
    \end{subfigure}
    \caption{Case 1. Dense-scene (69.9\% ROI) qualitative comparison showing the Input Image, Proposer Detections, Merged ROI Regions, ROI-Gated SAHI result, and Full SAHI baseline. ROI-gated processing yields near-baseline latency due to limited tile-saving potential.}
    \label{fig:qual7}
\end{figure*}

\begin{figure*}[!t]
\centering

\begin{subfigure}[t]{0.19\textwidth}
    \centering
    \includegraphics[width=\linewidth]{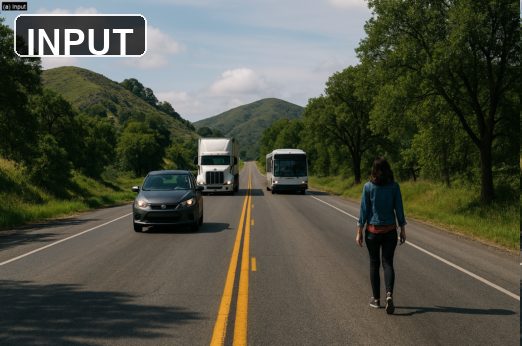}
    \caption{Input}
    \label{fig:qual2_input}
\end{subfigure}
\hfill
\begin{subfigure}[t]{0.19\textwidth}
    \centering
    \includegraphics[width=\linewidth]{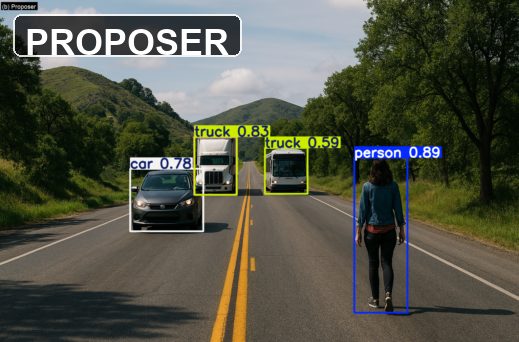}
    \caption{Proposer}
    \label{fig:qual2_proposer}
\end{subfigure}
\hfill
\begin{subfigure}[t]{0.19\textwidth}
    \centering
    \includegraphics[width=\linewidth]{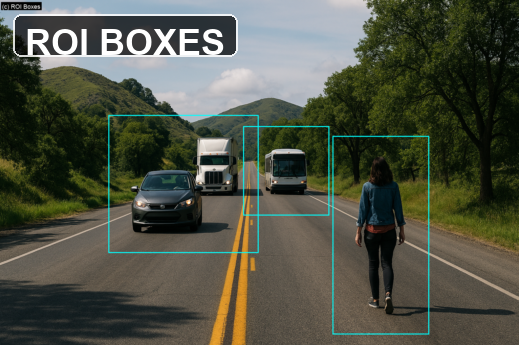}
    \caption{ROI Boxes}
    \label{fig:qual2_roiboxes}
\end{subfigure}
\hfill
\begin{subfigure}[t]{0.19\textwidth}
    \centering
    \includegraphics[width=\linewidth]{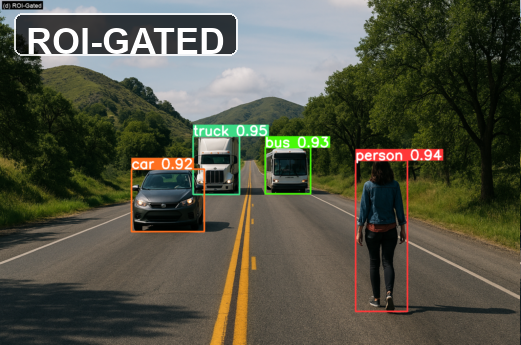}
    \caption{ROI-Gated}
    \label{fig:qual2_roigated}
\end{subfigure}
\hfill
\begin{subfigure}[t]{0.19\textwidth}
    \centering
    \includegraphics[width=\linewidth]{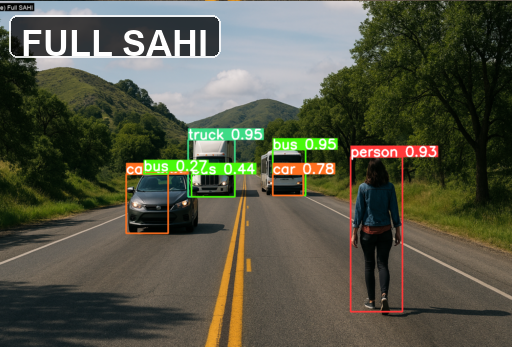}
    \caption{Full SAHI}
    \label{fig:qual2_fullsah}
\end{subfigure}
\caption{Case 2. The moderate-density (26.4\% ROI), where selective slicing reduces computation while preserving detection localization quality.}
\label{fig:qual2}

\end{figure*}

\begin{figure*}[!t]
\centering
\begin{subfigure} [t]{0.2\textwidth}
\includegraphics[width=\linewidth]{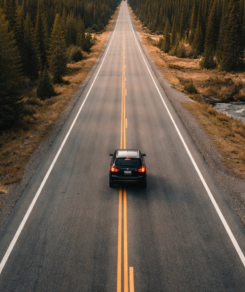}
    \caption{Input}
    \label{fig:qual3_a}
\end{subfigure}
\hfill
\begin{subfigure} [t]{0.165\textwidth}
\includegraphics[width=\linewidth]{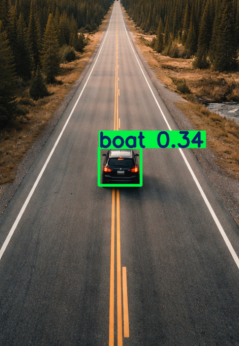}
    \caption{Proposer}
    \label{fig:qual3_b}
\end{subfigure}
\hfill
\begin{subfigure} [t]{0.156\textwidth}
\includegraphics[width=\linewidth]{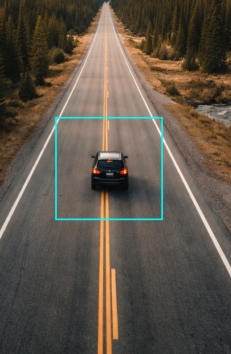}
    \caption{ROI boxes}
    \label{fig:qual3_c}
\end{subfigure}
\hfill
\begin{subfigure} [t]{0.167\textwidth}
\includegraphics[width=\linewidth]{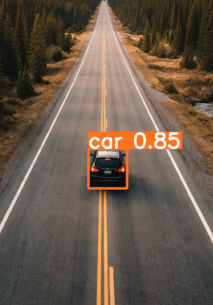}
    \caption{ROI gated}
    \label{fig:qual3_d}
\end{subfigure}
\hfill
\begin{subfigure} [t]{0.1975\textwidth}
\includegraphics[width=\linewidth]{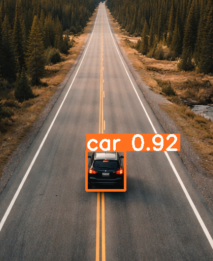}
    \caption{ROI SAHI}
    \label{fig:qual3_e}
\end{subfigure}
\caption{Case 3. Sparse-scene case (2.7\% ROI), where skipping background regions produces the largest efficiency gain.}
\label{fig:qual3}
\end{figure*}

\begin{figure*}[tp]
    \centering
\begin{subfigure}[t]{0.18\textwidth}
\centering
\includegraphics[width=\linewidth]{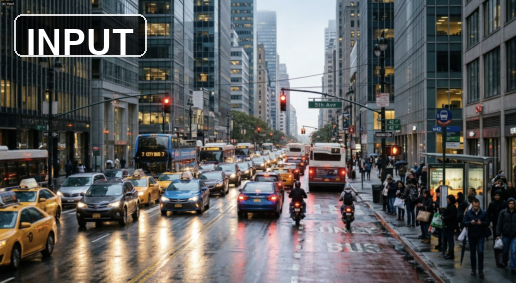}
\caption{Input}
\label{fig:qual5_input}
\end{subfigure}\hfill
\begin{subfigure}[t]{0.18\textwidth}
\centering
\includegraphics[width=\linewidth]{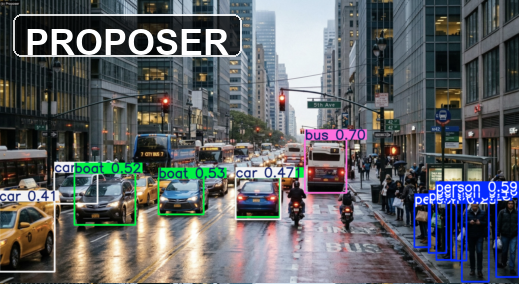}
\caption{Proposer}
\label{fig:qual5_proposer}
\end{subfigure}\hfill
\begin{subfigure}[t]{0.18\textwidth}
\centering
\includegraphics[width=\linewidth]{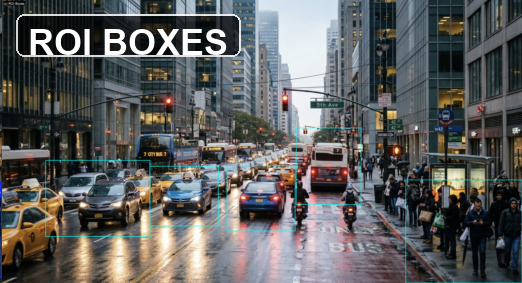}
\caption{ROI Regions}
\label{fig:qual5_roi}
\end{subfigure}\hfill
\begin{subfigure}[t]{0.18\textwidth}
\centering
\includegraphics[width=\linewidth]{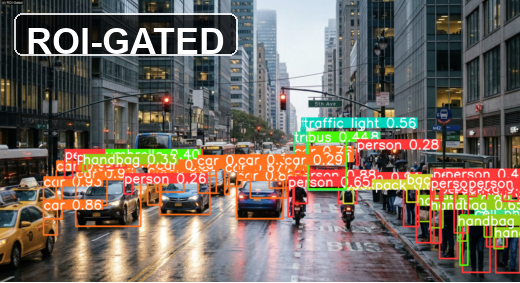}
\caption{ROI-Gated}
\label{fig:qual5_roi_gated}
\end{subfigure}\hfill
\begin{subfigure}[t]{0.18\textwidth}
\centering
\includegraphics[width=\linewidth]{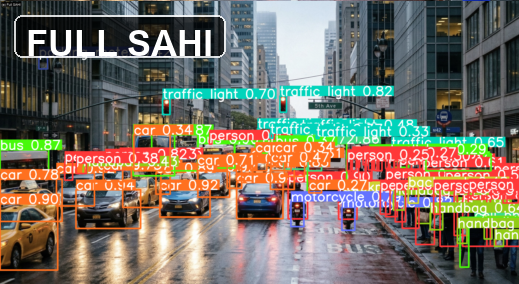}
\caption{Full SAHI}
\label{fig:qual5_full_sahi}
\end{subfigure}
    \caption{Case 2. Dense-scene qualitative comparison showing the Input Image, Proposer Detections, Merged ROI Regions, ROI-Gated SAHI result, and Full SAHI baseline.}
    \label{fig:qual5}
\end{figure*}

\begin{figure*}[!t]
\centering

\begin{subfigure}[t]{0.19\textwidth}
\centering
\includegraphics[width=\linewidth]{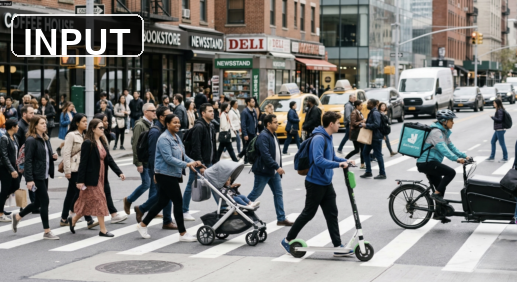}
\caption{Input}
\label{fig:qual4_input}
\end{subfigure}
\hfill
\begin{subfigure}[t]{0.19\textwidth}
\centering
\includegraphics[width=\linewidth]{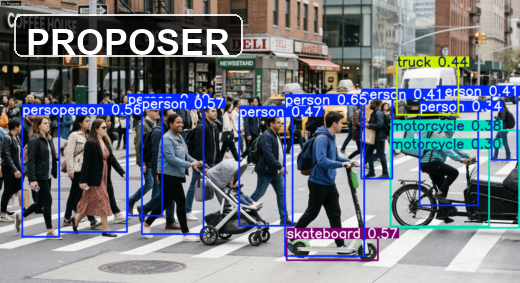}
\caption{Proposer}
\label{fig:qual4_proposer}
\end{subfigure}
\hfill
\begin{subfigure}[t]{0.19\textwidth}
\centering
\includegraphics[width=\linewidth]{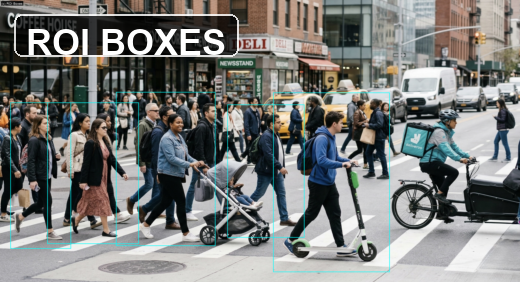}
\caption{Merged ROI Regions}
\label{fig:qual4_roi}
\end{subfigure}
\hfill
\begin{subfigure}[t]{0.19\textwidth}
\centering
\includegraphics[width=\linewidth]{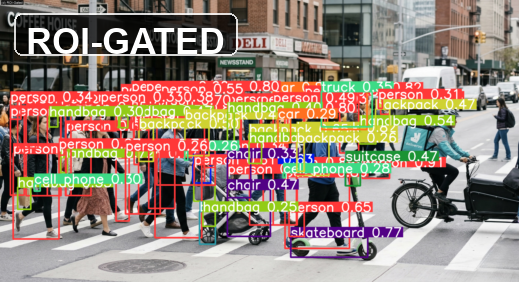}
\caption{ROI-Gated SAHI}
\label{fig:qual4_roi_gated}
\end{subfigure}
\hfill
\begin{subfigure}[t]{0.19\textwidth}
\centering
\includegraphics[width=\linewidth]{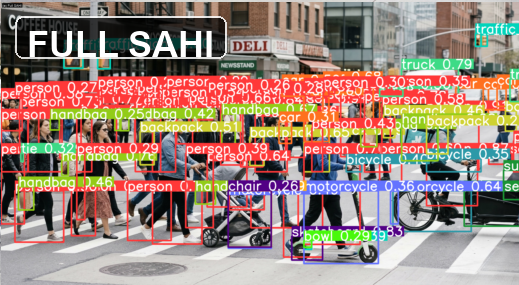}
\caption{Full SAHI}
\label{fig:qual4}
\end{subfigure}

\caption{Case 1. Dense-scene qualitative comparison showing the input image, proposer detections, merged ROI regions, ROI-Gated SAHI result, and Full SAHI baseline.}
\label{fig:qual4}
\end{figure*}

\begin{figure*}[!t]
    \centering

    \begin{subfigure}[b]{0.19\textwidth}
        \centering
        \includegraphics[width=\linewidth]{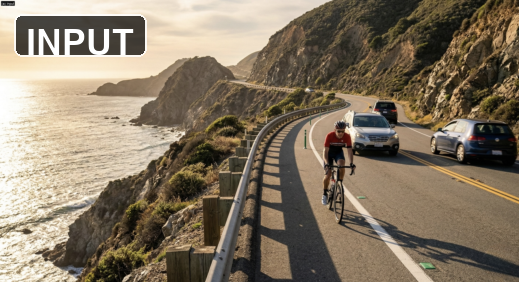}
        \caption{Input}
        \label{fig:qual6_input}
    \end{subfigure}
    \hfill
    \begin{subfigure}[b]{0.19\textwidth}
        \centering
        \includegraphics[width=\linewidth]{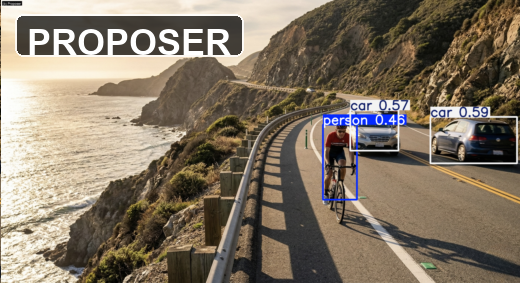}
        \caption{Proposer Detections}
        \label{fig:qual6_proposer}
    \end{subfigure}
    \hfill
    \begin{subfigure}[b]{0.19\textwidth}
        \centering
        \includegraphics[width=\linewidth]{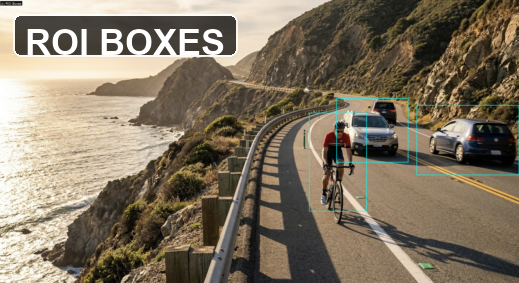}
        \caption{Merged ROI Regions}
        \label{fig:qual6_roi}
    \end{subfigure}
\hfill
    \begin{subfigure}[b]{0.19\textwidth}
        \centering
        \includegraphics[width=\linewidth]{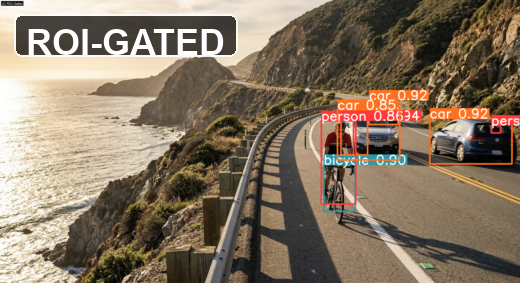}
        \caption{ROI-Gated SAHI}
        \label{fig:qual6_roi_gated}
    \end{subfigure}
    \hfill
    \begin{subfigure}[b]{0.19\textwidth}
        \centering
        \includegraphics[width=\linewidth]{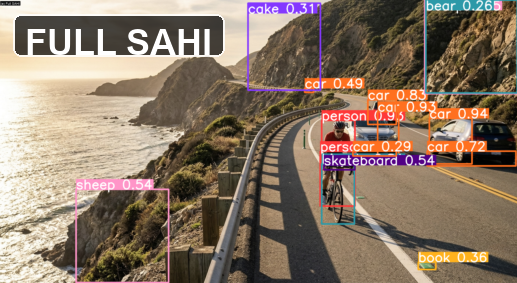}
        \caption{Full SAHI Baseline}
        \label{fig:qual6_full_sahi}
    \end{subfigure}
    \caption{Case 2. Moderate-scene qualitative comparison showing the Input Image, Proposer Detections, Merged ROI Regions, ROI-Gated SAHI result, and Full SAHI baseline.}
    \label{fig:qual6}
\end{figure*}

\textcolor{red}{Qualitative results demonstrate that ROI-Gated SAHI focuses computation on relevant regions while preserving spatial localization quality, as depicted in Figures~\ref{fig:qual7}--\ref{fig:qual6}.
The sparse-scene efficiency gain is consistent with Fig.~\ref{fig:latency} and Table~\ref{tab:per_image_speed}, where the 2.7\% ROI case improves from 526.52~ms to 76.25~ms ($6.90\times$ speedup).}


\subsection{Discussion}

\noindent\textbf{Agreement Metrics.} Agreement F1 measures the consistency between Full SAHI and ROI-Gated outputs in terms of detected objects, capturing how often both methods identify the same instances (including missed or additional detections). Agreement Mean IoU, on the other hand, evaluates the spatial alignment of matched detections by quantifying the overlap in bounding box position and size. High agreement means that ROI-Gated SAHI produces results similar to those of Full SAHI. Lower agreement indicates differences resulting from missed detections or changes in object localization. It is worth noting that these metrics compare the two methods with each other. Hence, these metrics do not measure absolute detection performance.

\subsubsection*{Impact of ROI Coverage}

The results show three distinct performance regimes based on coverage. In sparse scenes where the coverage is $<30\%$, the proposed method achieves the highest speedups, reaching up to $6.90\times$. This is because only a small part of the image requires processing.

In moderately dense scenes where the coverage is between $30\%$ and $75\%$, the method still provide stable performance gains, typically between $2$ times and $3\times$.
However, the results become more variable as the fraction of image requiring processing increases.
In dense scenes with coverage $>75\%$, the benefits of ROI-gating considerably decreases because most of the image requires processing. Consequently, the effectiveness of selective computation decreases.
This regime-dependent trend appears consistently in both the aggregate statistics and the case study results.

\subsubsection*{Efficiency--Accuracy Trade-off}

ROI-Gated SAHI reduces computation by  requiring fewer tiles to be processed. 
However, this computation improvement also affects detection accuracy.
The main reason is that the lightweight proposer may fail to identify some object regions. On the COCO128 full split, the mAP@0.5 decreases from  0.7569 for Full SAHI to 0.6602 for ROI-Gated SAHI. This suggests that some objects objects or regions are missed/not covered and not passed to the refinement stage, thereby propagating error to the refinement stage. 
Nevertheless, for objects detected by both methods, the localization quality of the objects for both methods is closely matched. This is indicated by the close alignment in IoU-based metrics.

\subsubsection*{Role of Adaptive Fallback}

Static ROI-gating does not perform well, particularly when ROI coverage is high. 
To address this issue, the adaptive fallback mechanism is introduced, which dynamically routes high-coverage images to Full SAHI. 
With $\tau = 0.40$, the average speed ratio increases from $0.88\times$ for static ROI-gating to $1.02\times$ relative to Full SAHI.
As a result, average-case slowdowns is removed while the the performance gain of ROI-Gating is retained for sparse scenes.

\subsubsection*{Practical Implications}

The proposed approach is suitable for  applications where scene density varies and low latency is critical. 
Particularly, in sparse scenes, the approach can achieve considerable computational saving without reducing detection quality. 
For datasets with both sparse and dense scenes, 
adaptive or policy-based routing should be considered to maintain stable performance. By dynamically selecting the most appropriate strategy for each image, such adaptive routing helps avoid regressions associated with static configurations.

\subsubsection*{Limitations}

The proposed framework relies on the recall of the lightweight proposer,. 
Missed regions at the proposal stage cannot be recovered in the later stages. Additionally, the 
framework uses fixed thresholds, such as ROI coverage threshold and ROI expansion ratio. These configurations may limit its generalizability, affecting its performance on datasets with different scene characteristics. 
Moreover, the given evaluations only consider
YOLO-based detectors, and does not include alternative detection architectures to validate the generalizability of the proposed framework.

\medskip
\noindent Overall, ROI-Gated SAHI 
reduces computational cost by limiting processing to selected regions of an image. At the same time, it maintains competitive detection performance. 
When integrated with adaptive routing, the proposed framework effectively operate under different scene densities. This characteristics make the framework a practical option to resource-constrained applications.

\section{Conclusion}
\label{sec:conclusion}


In this paper, we proposed ROI-Gated SAHI, a content-adaptive inference framework for computational-efficient slicing-based object detection. 
The framework introduces ROI proposer that estimates foreground distribution in an image before selectively applying sliced inference to those regions. This way, the framework reduces the redundant computations required in the conventional full-image SAHI where the entire image is processed, while preserving detection robustness across varying scene densities. 
For dense scenes, the framework uses an adaptive fallback mechanism, wherein processing is switched to full SAHI when ROI coverage becomes high. 
Experimental evaluation shows that on the COCO128 full split, static ROI-gating is slower than Full SAHI on average, achieving a high speed ratio of 0.88$\times$, and producing a lower mAP@0.5, decreasing from 0.6602 to 0.7569. 
The results also show that 
adaptive routing further improves this trend. Using a threshold of $\tau=0.40$, the average speed ratio increases to 1.02$\times$. This suggests that policy-based adaptive routing can increase average performance. 
Aside from this, a case study is conducted using three images with sparse, moderate, and dense ROI coverage, in which the speedups is observed between 0.96$\times$ and 6.90$\times$ with a mean value of  mean 3.41$\times$. 
The largest gains occur in sparse scenes, where only a small fraction of the image requires processing.
These results show that ROI-gated slicing is most effective when scene content is sparse. They also indicate that adaptive routing is important for maintaining stable performance across different scene conditions.
Future work will investigate adaptive and learned gating strategies. Additional directions include multi-scale ROI selection, class-aware ROI reasoning, and the application of ROI-gated inference to video data. These extensions may further improve efficiency and robustness in practical deployment scenarios.


\bibliographystyle{ieeetr}
\bibliography{references}






\end{document}